\documentclass[times, review, 10p]{elsarticle}

\usepackage{amssymb}
\usepackage{amsmath}
\usepackage{float}
\usepackage{wrapfig}
\usepackage{bbm}
\usepackage{etoolbox}
\usepackage{booktabs}
\usepackage{xcolor}

\journal{Pattern Recognition}

\begin{document}

\begin{frontmatter}

\title{SRAGAN: Saliency Regularized and Attended Generative Adversarial Network for Chinese Ink-wash Painting Style Transfer}

\author{Xiang Gao,Yuqi Zhang}{} 

\affiliation[Xiang Gao]{organization={Wangxuan Institute of Computer Technology, Peking University},
            addressline={No. 128 Zhongguancun North Street, Haidian District}, 
            city={Beijing},
            postcode={100871}, 
            country={China}}
            
\affiliation[Yuqi Zhang]{organization={School of Mathematical Sciences, University of Chinese Academy of Sciences},
            addressline={19A Yuquan Road, Shijingshan District}, 
            city={Beijing},
            postcode={100049}, 
            country={China}}

\makeatletter
\def\ps@pprintTitle{%
  \let\@oddhead\@empty
  \let\@evenhead\@empty
  \let\@oddfoot\@empty
  \let\@evenfoot\@oddfoot
}
\makeatother

\begin{abstract}
Recent style transfer problems are still largely dominated by Generative Adversarial Network (GAN) from the perspective of cross-domain image-to-image (I2I) translation, where the pivotal issue is to learn and transfer target-domain style patterns onto source-domain content images. This paper handles the problem of translating real pictures into traditional Chinese ink-wash paintings, i.e., Chinese ink-wash painting style transfer. Though a wide range of I2I models tackle this problem, a notable challenge is that the content details of the source image could be easily erased or corrupted due to the transfer of ink-wash style elements. To remedy this issue, we propose to incorporate saliency detection into the unpaired I2I framework to regularize image content, where the detected saliency map is utilized from two aspects: (\romannumeral1) we propose saliency IOU (SIOU) loss to explicitly regularize object content structure by enforcing saliency consistency before and after image stylization; (\romannumeral2) we propose saliency adaptive normalization (SANorm) which implicitly enhances object structure integrity of the generated paintings by dynamically injecting image saliency information into the generator to guide stylization process. Besides, we also propose saliency attended discriminator which harnesses image saliency information to focus generative adversarial attention onto the drawn objects, contributing to generating more vivid and delicate brush strokes and ink-wash textures. Extensive qualitative and quantitative experiments demonstrate superiority of our approach over related advanced image stylization methods in both GAN and diffusion model paradigms.
\end{abstract}



\begin{keyword}
Style Transfer, Image-to-Image Translation
\end{keyword}

\end{frontmatter}
  
\section{Introduction}
``What I cannot create, I do not understand.'' In recent years, advances in pattern recognition have witnessed transition from discriminative learning \cite{bib1}\cite{bib2} to generative machine intelligence \cite{bib3}\cite{bib4}, which imposes higher demands on the perception and understanding of underlying data patterns. The rapid progress of deep generative models has been revolutionizing various fields of generative AI, among which there are also artistic ones, leading to a growing prosperity of AI art creation tools and services, behind which neural style transfer (NST) \cite{bib5}, generative adversarial network (GAN) \cite{bib6}, and diffusion model \cite{bib7} are the main force of related technologies.

The seminal work of NST proposed by Gatys et al. \cite{bib5} for the first time utilizes convolutional neural network to separate and recombine image content and style, impressively reproducing artistic style patterns onto natural images. Shortly afterward, extensive efforts have been made to improve NST in visual quality \cite{bib8}, inference speed \cite{bib9}, and extension to multiple \cite{bib10} or arbitrary styles \cite{bib11}. Despite significant progress achieved, NST-related methods resort to texture-descriptor-based optimization functions to extract and represent style patterns, e.g., Gram loss \cite{bib5}, MMD loss \cite{bib12}, Mean-Variance loss \cite{bib13}, EMD loss \cite{bib14}, etc., which essentially transfer low-level texture patterns, not sufficient to capture more abstract high-level style distribution. 

By virtue of the inherent advantages of learning high-level data patterns, GAN is broadly applied to artistic stylization tasks, such as painting styliation \cite{bib15}, cartoon stylization \cite{bib16}, font stylization \cite{bib17}, etc., where the kernel ingredients behind these applications are the conditional GAN framework pioneered by Pix2Pix \cite{bib18} for supervised cross-domain I2I translation, or the bidirectional I2I frameworks represented by CycleGAN \cite{bib19} for unsupervised I2I translation. Different from NST-based methods using explicit texture losses to model image styles, GAN-based methods implicitly recognize and transfer target-domain style patterns with adversarial learning, bypassing the need to define specific style losses, all while possessing superiority in perceiving high-level style patterns.

\begin{figure}[t]
  \centering
  \includegraphics[width=\textwidth]{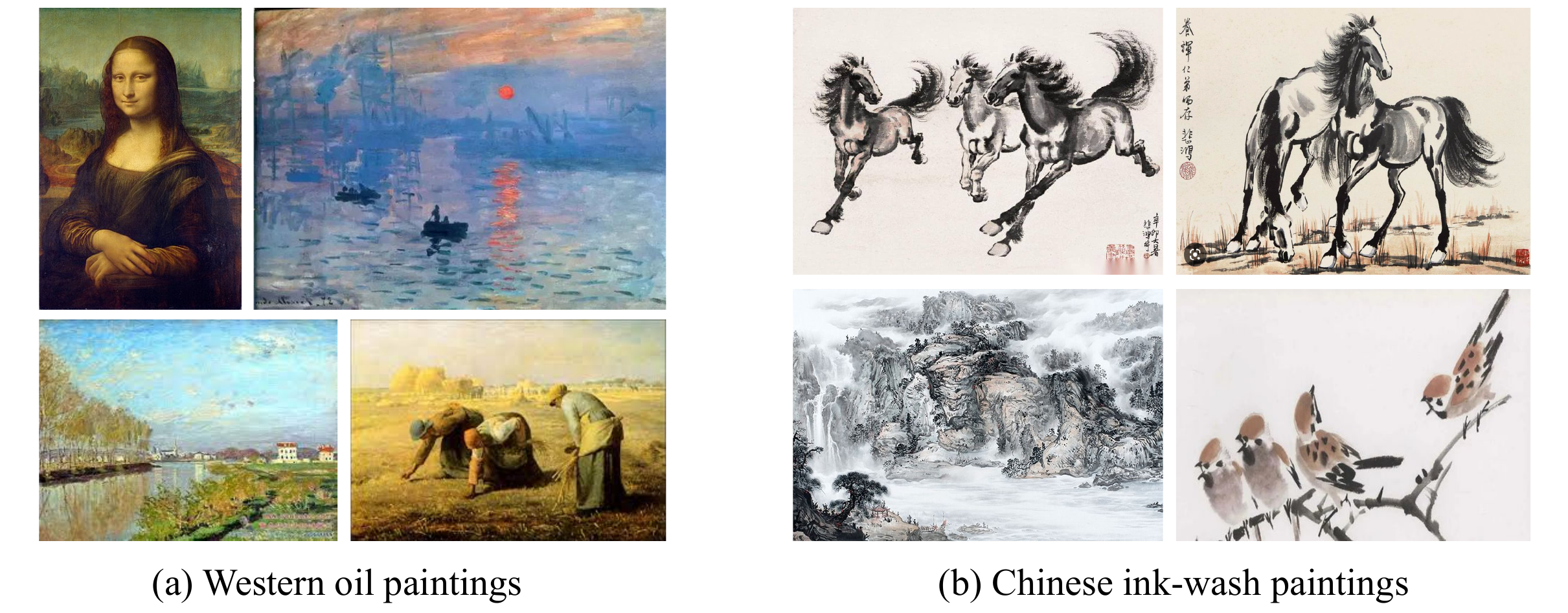}  
  \caption{Comparison between Western oil paintings and traditional Chinese ink-wash paintings.}
  \label{western_eastern_paintings}
\end{figure}

The advent of DDPM \cite{bib7} brings diffusion model significant attention in generative AI. Benefit from progress in multimodal representation learning and neural network architecture, large-scale text-to-image diffusion models achieve stunning success in text-guided artistic creation, among which Latent Diffusion Model (LDM) \cite{bib20} gains the most popularity by transferring diffusion model from pixel space to low-dimensional feature space to lower computational overhead. Based on LDM, a series of methods have been proposed to extend it from text-to-image generation to the realm of text-guided I2I translation \cite{bib21}\cite{bib22}\cite{bib23}\cite{bib24}\cite{bib25}. Different from GAN-based models that learn and transfer latent style patterns through adversarial training, these methods learn to encode image style patterns with textual embeddings by training multimodal diffusion models on massive image-text pairs, realizing open-domain image stylization instructed by arbitrary text prompts.

Among a variety of image stylization tasks, Chinese ink-wash painting style transfer is a relatively underexplored area. As shown in Fig. \ref{western_eastern_paintings}, as a representative art form of Chinese traditional culture, Chinese ink-wash painting differs from Western painting mainly in two aspects: (\romannumeral1) they are created by painting onto a canvas of paper or silk with a brush dipped with ink and water, rather than using pigments as Western paintings; (\romannumeral2) Chinese ink-wash paintings often feature landscapes, animals, birds, etc., they do not emphasize a precise depiction of scenes or objects, but instead the emotional statement of the painter reflected by the skills of applying ink. 

Efforts have been made to generate Chinese landscape paintings from random noise based on GAN \cite{bib26} or diffusion model \cite{bib27}. However, converting natural images into Chinese ink-wash paintings in an I2I paradigm is a more challenging task that requires both precise style rendering and faithful content preservation, especially in the case of unpaired training data. Moreover, the special characteristics of Chinese ink-wash paintings tend to make the discriminator network easily fooled with just irregular ink-wash style elements, regardless of content reasonableness, and thus leads to noticeable content missing, distortion, and abnormality after ink-wash stylization when directly applying general unsupervised I2I translation methods to this task.

To ameliorate this issue, ChipGAN \cite{bib28} introduces edge detection to the unsupervised I2I framework for Chinese ink-wash painting style transfer. It enforces consistency of the detected object contours before and after image stylization to strengthen object content integrity of the generated paintings. For the same purpose of content regularization, Chip-SAGAN \cite{bib29} proposes an edge-promoting adversarial loss to promote object contours of the translated Chinese ink-wash paintings. 

Nevertheless, we argue that enforcing object contour consistency may not be the optimal solution to regularize object content structure. Firstly, Chinese ink-wash paintings do not emphasize detailed depiction of objects, some object edges in the drawn paintings can be reasonably omitted to better cater to the corresponding drawing skills. Secondly, Chinese ink-wash painting is a highly abstract artistic form, the drawn object contours should not be strictly aligned with the source image contours in pixel level. These inherent painting characteristics contradict with the rationality of contour consistency loss. Furthermore, regions inside object contours also count for object structure, whereas contour-based objective functions are not able to regularize such internal structural information.

Therefore, we propose to combine unsupervised cross-domain I2I translation with saliency detection for content-regularized Chinese ink-wash painting style transfer. Our motivations to focus on image saliency are fourfolds: (\romannumeral1) Chinese ink-wash paintings draw scenes or objects with dark brush strokes on white canvas, the drawn objects are prominent and the background is blank, which is suitable for saliency detection to segment out the drawn objects. (\romannumeral2) The detected saliency map highlights object location, shape, contour, and pixel intensity, and thus contains richer structural information than edges when used for content regularization. (\romannumeral3) It is intuitively reasonable that the salient object in real photos should also be visually salient in the translated ink-wash paintings.
(\romannumeral4) The detected saliency map provides an informative instruction to where the adversarial learning of the target style patterns should be focused on, such that detailed stylization quality of the drawn objects could be improved as compared with evenly attending to all image pixels.

\begin{figure}[t]
  \centering
  \includegraphics[width=\textwidth]{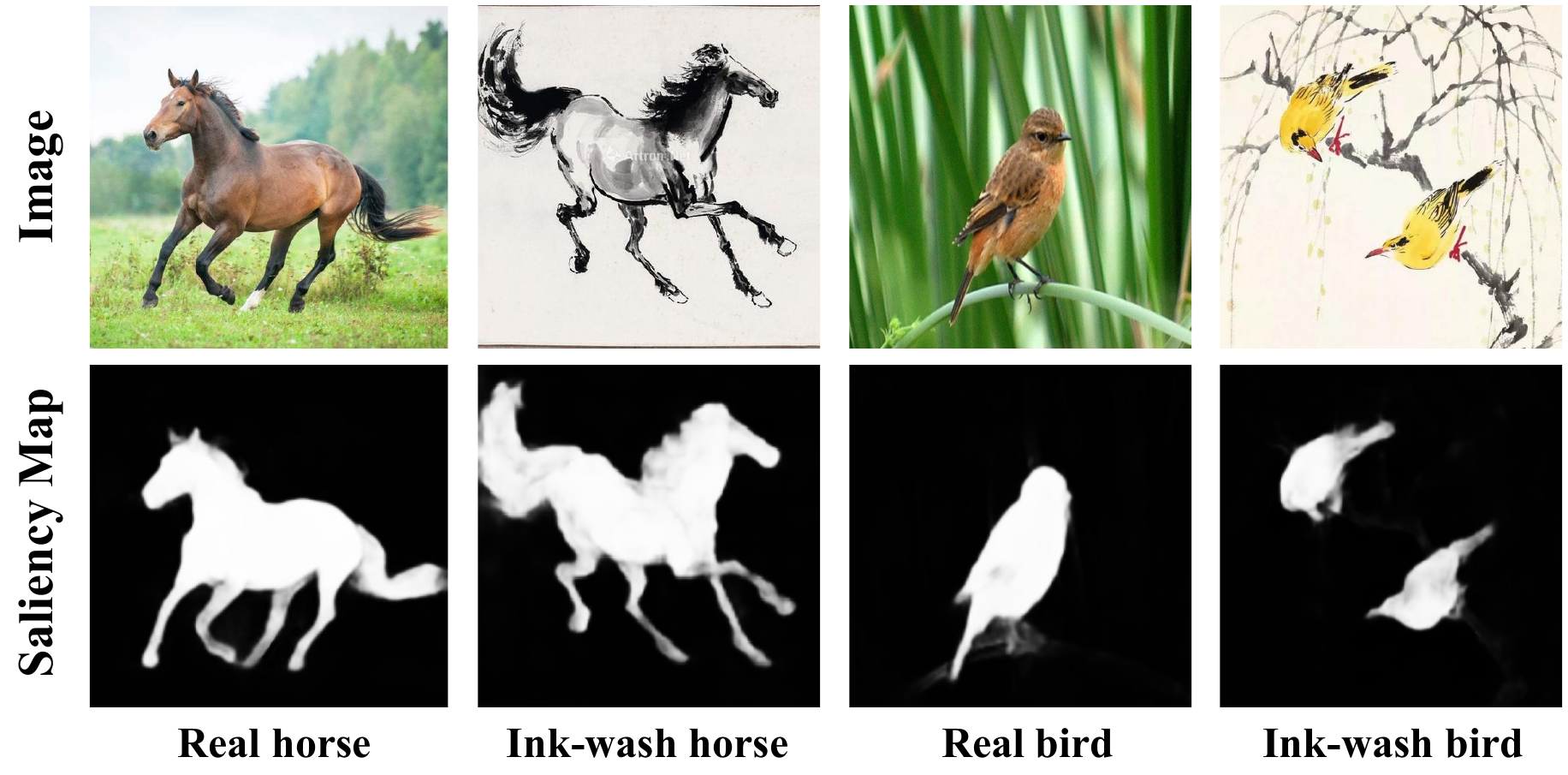}  
  \caption{Top: real-world images and Chinese ink-wash paintings. Bottom: saliency maps detected by CSNet \cite{bib30}. The saliency detection deep model pre-trained on real-world images can also segment out salient objects well for Chinese ink-wash paintings.}
  \label{saliency_visualize}
\end{figure}

However, there comes a question: is the saliency detection model pre-trained on real-world images also applicable to the domain of Chinese ink-wash paintings? The answer is yes. We evaluate pre-trained saliency detection model CSNet \cite{bib30} on both two domains, example results displayed in Fig. \ref{saliency_visualize} demonstrate reliable generalization ability: the saliency detection deep model trained on real-world-image manifold can also segment out salient objects well on Chinese ink-wash paintings. Based on these motivations and observations, we propose saliency regularized and attended generative adversarial network (SRAGAN) for Chinese ink-wash painting style transfer. Established on the unsupervised I2I translation GAN framework, our method leverages saliency detection to regularize object structure in the translated ink-wash paintings, alleviating content missing and distortion issues caused by transfer of irregular ink-wash style elements. To penalize object structure distortion, we propose a saliency-constrained learning objective by constraining consistency between the detected saliency maps before and after image stylization. Considering that Chinese ink-wash painting encourages mild object contour displacement to cater to abstractness and typical brush stroke styles, we do not enforce strict per-pixel consistency of saliency maps but instead a relaxed saliency consistency loss that maximizes IOU of the thresholded saliency maps before and after stylization, which we term saliency IOU (SIOU) loss. Our proposed SIOU loss regularizes object structure of the generated paintings without degrading style rendering quality. In addition, we devise a saliency regularized generator network. It utilizes the detected saliency map as structural guidance and dynamically integrates its structural information into the feed-forward process via our proposed saliency adaptive normalization (SANorm) layer, which implicitly regularizes object content structure by enhancing network's structure preservation capability. Furthermore, we harness image saliency information to adaptively guide adversarial learning attention onto salient image objects, such that transfer of the target style patterns can be specifically focused on the drawn objects to promote detailed stylization quality. It is worth mentioning that though diffusion model has outperformed GAN in image synthesis, its training paradigm makes it less flexible to separately model image content and style for I2I style transfer tasks. We demonstrate in the experiment section that our SRAGAN can also produce more satisfying Chinese ink-wash paintings than existing advanced diffusion based I2I stylization methods.

To sum up, we propose a GAN-based Chinese ink-wash painting style transfer method that leverages saliency detection to improve object structure integrity and stylization quality of the drawn objects, achieving leading stylization effects in generation of different types of Chinese ink-wash paintings among related advanced methods. The kernel contributions are summarized as follows:

\begin{itemize}
	\item We propose saliency IOU (SIOU) loss which alleviates content corruption issue of Chinese ink-wash painting style transfer by explicitly regularizing object saliency consistency in a relaxed IOU-based constraint.
	\item We propose saliency regularized generator which implicitly enhances network's object structure preservation capability by dynamically integrating image saliency information into the feed-forward process via our proposed saliency adaptive normalization (SANorm).
	\item We propose saliency attended discriminator which contributes to producing more delicate and vivid ink-wash brush strokes and textures by utilizing saliency information to guide generative adversarial learning attention specifically onto the drawn objects.
\end{itemize}

\section{Related Work}
\subsection{Style Transfer}
Style transfer techniques have made rapid progress in recent years, where a general solution is to employ adversarial learning to transfer target-domain style patterns while using perceptual loss \cite{bib9} or cycle-consistency loss \cite{bib19} to maintain original image content. Advanced style transfer methods focus on real-time transfer of arbitrary styles, where the kernel ingredient lies in the dynamical fusion of content features and style features during network forward propagation, as represented by the classical AdaIN layer \cite{bib13} and WCT layer \cite{bib31}. More advanced methods such as SANet \cite{bib32} and Adaattn \cite{bib33} further enrich the transferred style patterns by resorting to self-attention for feature fusion. More recently, diffusion-based style transfer methods receive plenty of attention. Stylediffusion \cite{bib34} combines CLIP and diffusion model to disentangle and recombine image content and style. Based on pre-trained large diffusion model, Freestyle \cite{bib24} proposes a training-free feature modulation approach to realize text-guided style transfer. In general, style transfer techniques are making continual progress with the development of generative AI.

\subsection{Unsupervised Image-to-image Translation}
Unsupervised I2I translation is especially catered for artistic authoring applications where paired training data are not available. These methods learn to transfer domain distributions with adversarial training while maintaining source-domain content information through cycle-consistency constraint (e.g., CycleGAN \cite{bib19}), shared latent space assumption (e.g., UNIT \cite{bib35}), self-supervised constraint (e.g., GcGAN \cite{bib36}), or contrastive learning (e.g., CUT \cite{bib37}). Recently, large-scale diffusion models have exhibited potential in open-domain I2I translation. These methods edit image styles using text prompts, and meanwhile preserving original content via null-text embedding optimization (e.g., Null-text Inversion \cite{bib22}), plug-and-play attention maps modulation (e.g., PnP \cite{bib23}), dynamic feature frequency band transplantation (e.g., FBSDiff \cite{bib25}), or intermediate noising followed by text-conditioned denoising (e.g., SDEdit \cite{bib21}).

\section{Method}
Our SRAGAN model is built upon a bidirectional GAN framework which establishes mutual translation mappings between source and target domains. Let ${\mathcal{X}}$ be the source domain of real-world images, ${\mathcal{Y}}$ be the target domain of Chinese ink-wash paintings (the image data of the two domains are unpaired), the model is trained with a randomly sampled image pair $\{x \in \mathcal{X}, y \in \mathcal{Y}\}$ at each training iteration, and is inferred by feeding in a real-world picture to obtain the generated Chinese ink-wash painting via the learned source-to-target translation mapping ${\mathcal{X}} \rightarrow {\mathcal{Y}}$. 

\subsection{Overall Architecture}
\begin{figure}[t]
  \centering
  \includegraphics[width=\textwidth]{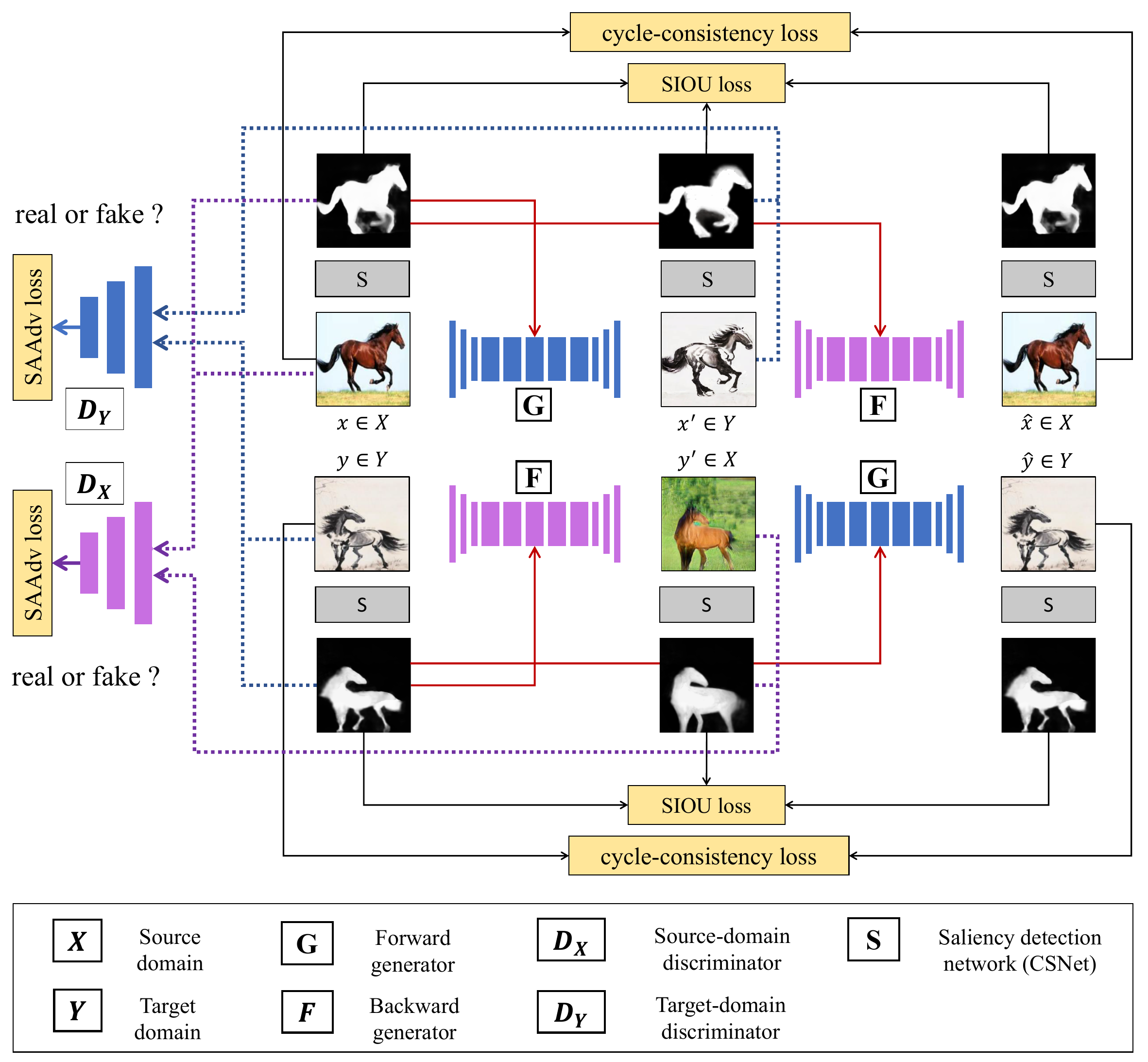}  
  \caption{Overall architecture of our SRAGAN model.} 
  \label{overall_architecture}
\end{figure}

As illustrated in Fig. \ref{overall_architecture}, the generator $G$ transforms a source-domain real image $x \in \mathcal{X}$ into a target-domain painting $x'=G(x) \in \mathcal{Y}$, the generator $F$ inversely learns to translate a target-domain painting $y \in \mathcal{Y}$ into a source-domain real image $y'=F(y) \in \mathcal{X}$. The target-domain discriminator $D_{\mathcal{Y}}$ acts as an adversary against the generator $G$ by discriminating synthesized paintings against real paintings. The source-domain discriminator $D_{\mathcal{X}}$ learns to distinguish the real-world images synthesized by $F$ from the real source-domain natural images. Meanwhile, cycle-consistency constraint is employed to avoid cross-domain translation content mismatch by enforcing mutual invertibility of the two generators, i.e., $x \approx \hat{x}=F(x')=F(G(x))$; $y \approx \hat{y}=G(y')=G(F(y))$. 

To alleviate content corruption issue and further improve image stylization quality, we incorporate saliency detection into the above-mentioned base model, proposing a saliency regularized and attended generative adversarial learning framework. We use CSNet \cite{bib30}, an efficient pre-trained saliency detection model denoted as $S$ to produce saliency maps that highlight regions of salient objects. The saliency information is utilized from three aspects: (1) we design a saliency regularized generator that leverages image saliency information to guide network forward propagation to enhance its object structure preservation ability; (2) we explicitly enforce saliency consistency before and after ink-wash stylization via our proposed saliency IOU (SIOU) loss; (3) we propose saliency attended discriminator which leverages image saliency information to focus ink-wash stylization attention onto the drawn objects for improved style rendering.

\begin{figure}[t]
  \centering
  \includegraphics[width=\textwidth]{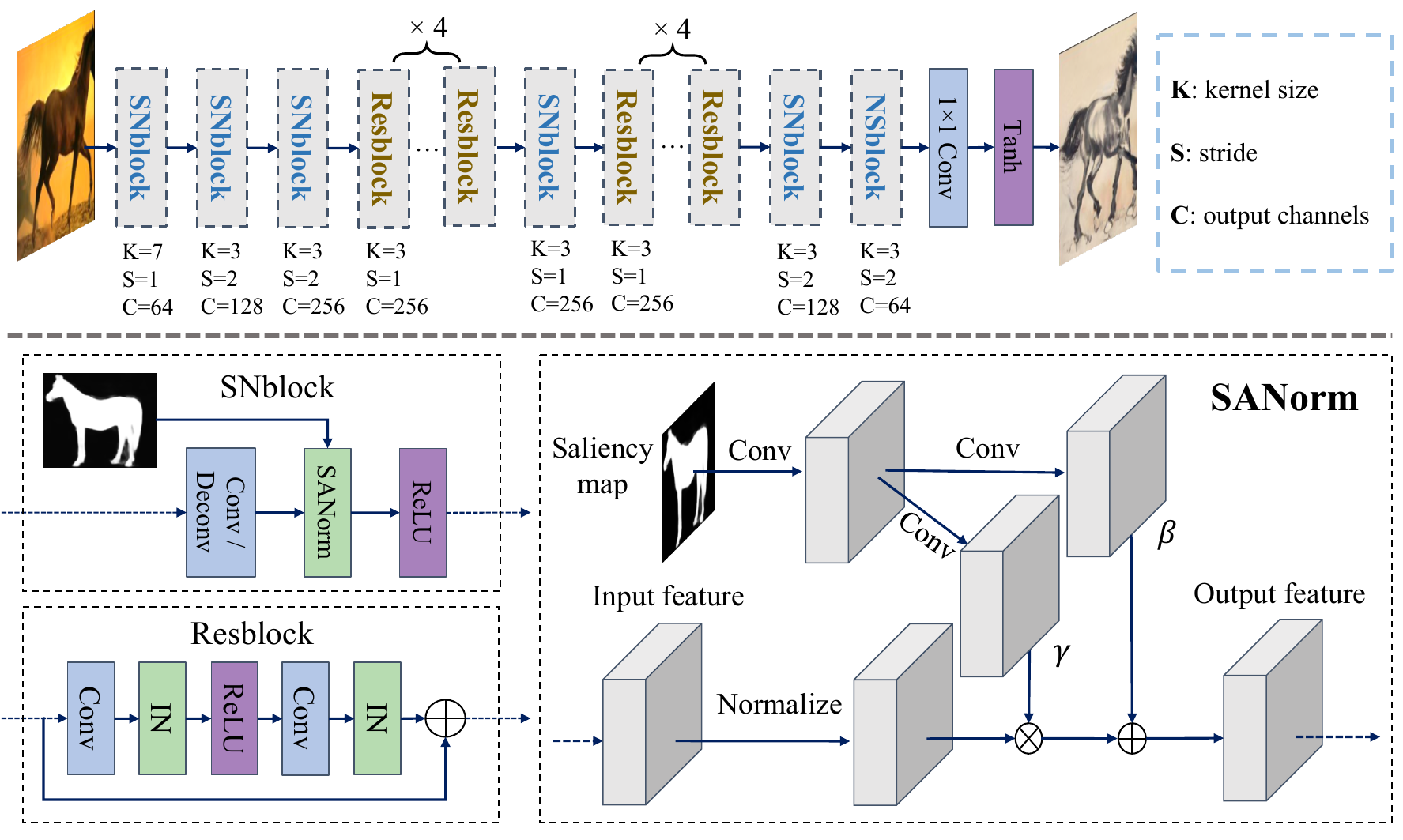}  
  \caption{Architecture details of our proposed saliency regularized generator.} 
  \label{generator_architecture}
\end{figure}

\subsection{Model Components}
\subsubsection{Saliency Regularized Generator}
The architectural details of our designed saliency regularized generator are illustrated in Fig. \ref{generator_architecture}. It follows a normal encoder-decoder structure with two types of basic convolutional units: residual convolutional block (Resblock) and saliency normalized convolutional block (SNblock). In all the convolutional units, instance normalization (IN) is used to accelerate learning of target-domain style patterns. The kernel ingredient of SNblock is our proposed saliency adaptive normalization (SANorm) layer. SANorm learns to adaptively fuse structural information of the detected saliency map into the current convolutional features to enhance object content structure preservation. Details of the SANorm layer are illustrated in Fig. \ref{generator_architecture}. Suppose the input feature of the SANorm layer is $f^{in}$ that has shape of $\{C, H, W\}$, where $C$ denotes channel number, $H$ and $W$ are feature height and width respectively, $f^{in}$ is firstly spatially normalized to the standard Gaussian distribution:
\begin{equation}
\tilde{f}^{in}_{c,h,w}=\frac{f^{in}_{c,h,w}-\mu_{c}}{\sigma_{c}},
\end{equation}
where $c$, $h$, $w$ are indices along the channel, height, and width dimension, $\mu_{c}$ and $\sigma_{c}$ are mean and standard deviation of the $c$th channel of $f^{in}$: 
\begin{equation}
\mu_{c}=\frac{1}{HW}\sum_{h,w}f^{in}_{c,h,w},\quad \sigma_{c}=\sqrt{\frac{1}{HW}\sum_{h,w}(f^{in}_{c,h,w}-\mu_{c})^{2}}.
\end{equation}
Then, the normalized feature $\tilde{f}^{in}$ is modulated with element-wise multiplication and addition to yield the output feature 
$f^{out}$:

\begin{equation}
f^{out}=\tilde{f}^{in} \times \gamma + \beta,
\end{equation}
where $\gamma$ and $\beta$ are respectively the scaling tensor and shifting tensor that has the same shape as $f^{in}$. The $\gamma$ and $\beta$ are learned from the input saliency map with two sibling convolutional branches $g_{1}$ and $g_{2}$:
\begin{equation}
\gamma=g_{1}(S), \quad \beta=g_{2}(S).
\end{equation}
Both $g_{1}$ and $g_{2}$ comprise two $3\times 3$ convolutions joined by $ReLU$, where parameters of the first convolution are shared to save computation overhead.

\subsubsection{Saliency Attended Discriminator}

\begin{figure}[t]
\centering
\includegraphics[width=\textwidth]{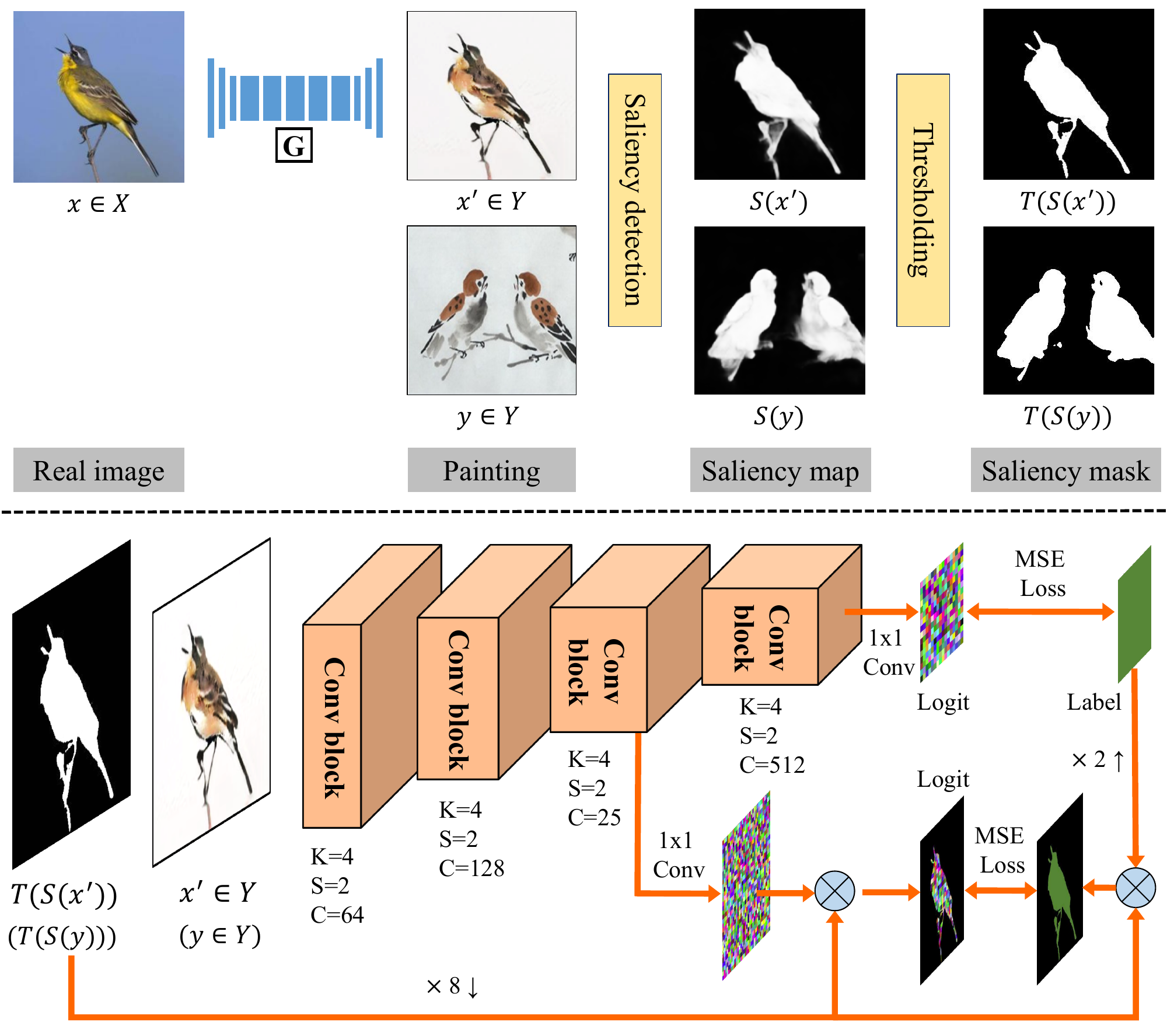}
\caption{Architecture details of our proposed saliency attended discriminator.}
\label{discriminator_architecture}
\end{figure}

To improve style rendering quality, we propose saliency attended discriminator to adaptively focus generative adversarial stylization attention onto salient image objects. As illustrated in Fig. \ref{discriminator_architecture}, taking the target-domain discriminator $D_{\mathcal{Y}}$ as an example, it basically follows PatchGAN \cite{bib18} architecture which outputs a matrix with each element indicating the authenticity score of a certain patch in the input image. The network is composed of four downsampling convolutional blocks, and comprises a main discrimination branch for global-level adversarial learning and a saliency-based auxiliary discrimination branch for fine-grained local-level adversarial learning. 

In the main discrimination branch, feature maps after the fourth convolutional block are converted to a $\frac{1}{16}$-scale single-channel logit matrix, which is drawn toward the corresponding $\frac{1}{16}$-scale label matrix with MSE loss.

In parallel to the main discrimination branch, a saliency-based auxiliary discrimination branch is appended to focally examine finer-grained patch authenticity over the object region. Specifically, a $\frac{1}{8}$-scale logit matrix derived from the third convolutional block as well as the corresponding $\frac{1}{8}$-scale label matrix are multiplied with a binary saliency mask which locates the salient objects (i.e., the drawn objects) in an image, The masked logit matrix is drawn toward the masked label matrix under the same MSE loss for local-level saliency attended adversarial learning. The binary saliency mask is synthesized by resizing the detected saliency map to the same $\frac{1}{8}$ scale followed by binary thresholding with 0.5 threshold (the value range of the detected saliency map is [0, 1]). By focusing finer-grained generative adversarial learning specifically onto the object region guided by image saliency information, the drawn object (horse, bird, etc.) can be better rendered with more delicate and vivid ink-wash brush strokes. 

\subsection{Objective Functions}
Our SRAGAN includes three loss functions: saliency attended adversarial loss $L_{SAAdv}$, cycle consistency loss $L_{cycle}$, and saliency IOU loss $L_{SIOU}$. Below are the formulations of these loss functions.

\subsubsection{Saliency Attended Adversarial Loss}
The saliency attended adversarial loss $L_{SAAdv}$ is composed of a generator part $L_{SAAdv\_G}$ and a discriminator part $L_{SAAdv\_D}$, both two parts can be further decomposed into a $\mathcal{X} \rightarrow \mathcal{Y}$ item and a $\mathcal{Y} \rightarrow \mathcal{X}$ one. 
\begin{equation}
L_{SAAdv} = L^{\mathcal{X} \rightarrow \mathcal{Y}}_{SAAdv\_G} + L^{\mathcal{Y} \rightarrow \mathcal{X}}_{SAAdv\_G} + L^{\mathcal{X} \rightarrow \mathcal{Y}}_{SAAdv\_D} + L^{\mathcal{Y} \rightarrow \mathcal{X}}_{SAAdv\_D}.
\end{equation}
Taking the $\mathcal{X} \rightarrow \mathcal{Y}$ mapping as example, let $D^{1}_{\mathcal{Y}}(\cdot)$ denote the output $\frac{1}{16}$-scale logit matrix at the main branch of $D_{\mathcal{Y}}$, $D^{2}_{\mathcal{Y}}(\cdot)$ denote the predicted $\frac{1}{8}$-scale logit matrix at the auxiliary branch, $\mathbbm{1}^{1}$ be an all-one matrix of the same spatial size as $D^{1}_{\mathcal{Y}}(\cdot)$, $\mathbbm{1}^{2}$ be an all-one matrix of the same size as $D^{2}_{\mathcal{Y}}(\cdot)$, $S(\cdot)$ and $T(\cdot)$ denote the saliency detection network and the binary thresholding operation respectively, $\downarrow$ denotes $\times8$ bilinear downsampling, then:
\begin{equation}
\begin{aligned}
L^{\mathcal{X} \rightarrow \mathcal{Y}}_{SAAdv\_D}=& \mathbbm{E}_{y \in \mathcal{Y}}[\frac{\sum{[(D^{1}_{\mathcal{Y}}(y)-\mathbbm{1}^{1})^{2}]}}{\sum{\mathbbm{1}^{1}}} + \frac{\sum{[T(S(y)_{\downarrow}) \otimes (D^{2}_{\mathcal{Y}}(y) - \mathbbm{1}^{2})^{2}]}}{\sum{[T(S(y)_{\downarrow}) \otimes \mathbbm{1}^{2}]}}] + \\
& \mathbbm{E}_{x'}[\frac{\sum{[D^{1}_{\mathcal{Y}}(x')^{2}]}}{\sum{\mathbbm{1}^{1}}} + \frac{\sum{[T(S(x')_{\downarrow}) \otimes D^{2}_{\mathcal{Y}}(
x')^{2}]}}{\sum{[T(S(x')_{\downarrow}) \otimes \mathbbm{1}^{2}]}}],
\end{aligned}
\end{equation}
\begin{equation}
\begin{aligned}
L^{\mathcal{X} \rightarrow \mathcal{Y}}_{SAAdv\_G}=& \mathbbm{E}_{x'}[\frac{\sum{[(D^{1}_{\mathcal{Y}}(x')-\mathbbm{1}^{1})^{2}]}}{\sum{\mathbbm{1}^{1}}} + \frac{\sum{[T(S(x')_{\downarrow}) \otimes (D^{2}_{\mathcal{Y}}(x') - \mathbbm{1}^{2})^{2}]}}{\sum{[T(S(x')_{\downarrow}) \otimes \mathbbm{1}^{2}]}}],
\end{aligned}
\end{equation}
where $x'$ is the generated painting by the saliency regularized generator $G$ given the input real-world image $x \in \mathcal{X}$, i.e., $x'=G(x, S(x))$, $\otimes$ denotes elementwise multiplication. Similarly, the losses for the inverse $\mathcal{Y} \rightarrow \mathcal{X}$ mapping is:
\begin{equation}
\begin{aligned}
L^{\mathcal{Y} \rightarrow \mathcal{X}}_{SAAdv\_D}=& \mathbbm{E}_{x \in \mathcal{X}}[\frac{\sum{[(D^{1}_{\mathcal{X}}(x)-\mathbbm{1}^{1})^{2}]}}{\sum{\mathbbm{1}^{1}}} + \frac{\sum{[T(S(x)_{\downarrow}) \otimes (D^{2}_{\mathcal{X}}(x) - \mathbbm{1}^{2})^{2}]}}{\sum{[T(S(x)_{\downarrow}) \otimes \mathbbm{1}^{2}]}}] + \\
& \mathbbm{E}_{y'}[\frac{\sum{[D^{1}_{\mathcal{X}}(y')^{2}]}}{\sum{\mathbbm{1}^{1}}} + \frac{\sum{[T(S(y')_{\downarrow}) \otimes D^{2}_{\mathcal{X}}(
y')^{2}]}}{\sum{[T(S(y')_{\downarrow}) \otimes \mathbbm{1}^{2}]}}],
\end{aligned}
\end{equation}
\begin{equation}
\begin{aligned}
L^{\mathcal{Y} \rightarrow \mathcal{X}}_{SAAdv\_G}=& \mathbbm{E}_{y'}[\frac{\sum{[(D^{1}_{\mathcal{X}}(y')-\mathbbm{1}^{1})^{2}]}}{\sum{\mathbbm{1}^{1}}} + \frac{\sum{[T(S(y')_{\downarrow}) \otimes (D^{2}_{\mathcal{X}}(y') - \mathbbm{1}^{2})^{2}]}}{\sum{[T(S(y')_{\downarrow}) \otimes \mathbbm{1}^{2}]}}],
\end{aligned}
\end{equation}
where $y'$ is the generated real image by the saliency regularized generator $F$ given the input real painting $y \in \mathcal{\mathcal{Y}}$, i.e., $y'=F(y, S(y))$. 

\subsubsection{Cycle Consistency Loss}
The cycle consistency loss $L_{cycle}$ enforces mutual invertibility of $G$ and $F$:
\begin{equation}
\begin{aligned}
L_{cycle}=& \mathbbm{E}_{x \in \mathcal{X}}[\|x - F(G(x, S(x)), S(x))\|_{1}] + \\
          & \mathbbm{E}_{y \in \mathcal{Y}}[\|y - G(F(y, S(y)), S(y))\|_{1}].
\end{aligned}
\end{equation}

\subsubsection{Saliency IOU Loss}
The saliency IOU loss $L_{SIOU}$ regularizes object saliency consistency before and after image stylization. It maximizes the IOU between the binary saliency mask of the generator input and output, as well as between the binary saliency mask of the generator input and its cyclically recovered result:
\begin{equation}
\begin{aligned}
L_{SSIOU}=& -\mathbbm{E}_{x \in \mathcal{X}}[\frac{T(S(x)) \cap T(S(x'))}{T(S(x)) \cup T(S(x'))}+       \frac{T(S(x)) \cap T(S(\hat{x}))}{T(S(x)) \cup T(S(\hat{x}))}] - \\
          & \mathbbm{E}_{y \in \mathcal{Y}}[\frac{T(S(y)) \cap T(S(y'))}{T(S(y)) \cup T(S(y'))}+       \frac{T(S(y)) \cap T(S(\hat{y}))}{T(S(y)) \cup T(S(\hat{y}))}],
\end{aligned}
\end{equation}
in which $x'$ is the painting generated by $G$ given the input real picture $x$, i.e., $x'=G(x,(S(x)))$, $\hat{x}$ is the recovered result of $x$ inverted by the backward generator $F$, i.e., $\hat{x}=F(G(x, S(x)), S(x))$. Likewise, $y'=F(y,(S(y)))$, $\hat{y}=G(F(y, S(y)), S(y))$. We resort to IOU-based saliency consistency constraint to enforce only overall object structure similarity, rather than strictly enforcing per-pixel consistency between saliency maps. Such relaxed structural constraint allows mild contour displacement in the stylized paintings and thus enables the network to focus on synthesizing more delicate and vivid artistic brush strokes and textures, producing style patterns that are more in line with the specific drawing skills of Chinese ink-wash painting. As visually verified in the experiment section, our IOU-based saliency consistency loss avoids rigid stylization effects caused by pixel-level saliency consistency constraint.

\subsubsection{Total Loss}
The total loss $L_{total}$ is the weighted combination of component losses:
\begin{equation}
L_{total}=\lambda_{1}L_{SAAdv} + \lambda_{2}L_{cycle} + \lambda_{3}L_{SIOU},
\label{total_loss}
\end{equation}
where $\lambda_{1}$, $\lambda_{2}$, and $\lambda_{3}$ are weights of the component loss functions.

\begin{figure}[t]
\centering
\includegraphics[width=0.94\textwidth]{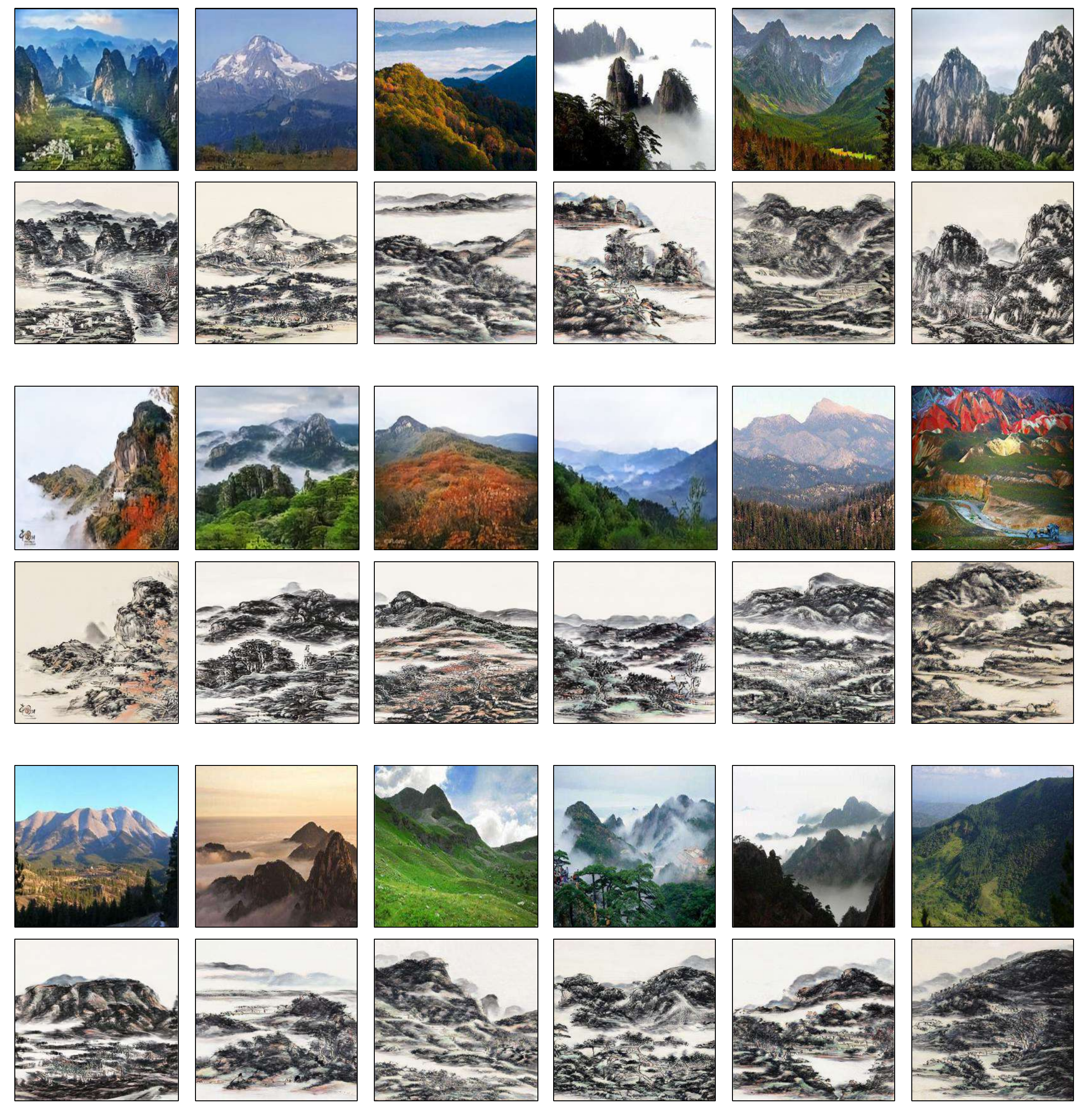}
\caption{Example results of our SRAGAN for converting real landscape photos (the top row) into Chinese ink-wash landscape paintings (the bottom row). Results are evaluated on the test set of the RealLandscape dataset. Better viewed with zoom-in.}
\label{landscape_res}
\end{figure}

\begin{figure}[t]
\centering
\includegraphics[width=\textwidth]{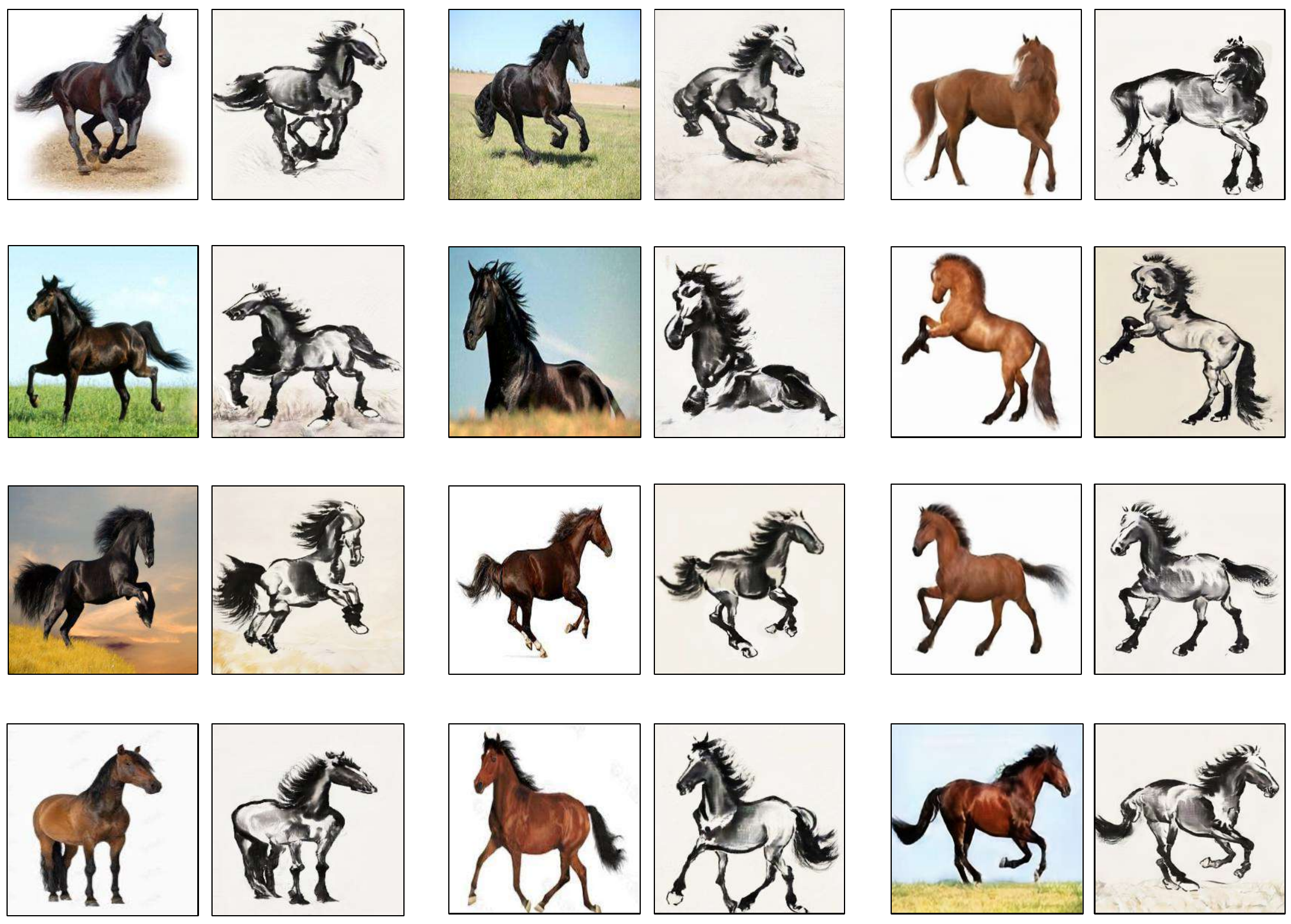}
\caption{Example results of our SRAGAN for converting real horse pictures (the left column) into Chinese ink-wash horse paintings (the right column). Results are evaluated on the test set of the RealHorses dataset. Better viewed with zoom-in.}
\label{horses_res}
\end{figure}

\begin{figure}[t]
\centering
\includegraphics[width=0.87\textwidth]{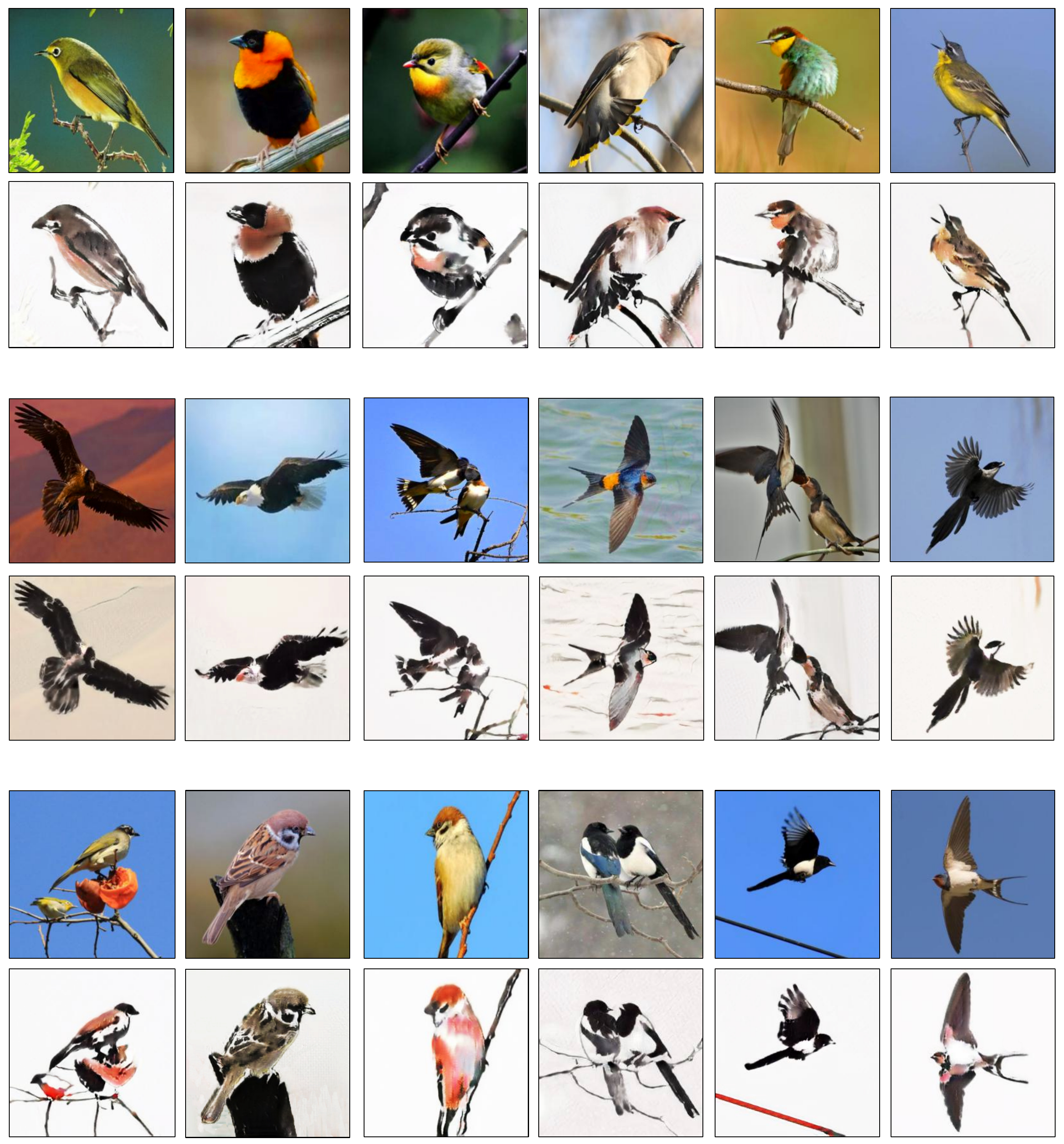}
\caption{Example results of our SRAGAN for converting real bird pictures (the top row) into Chinese ink-wash bird paintings (the bottom row). Results are evaluated on the test set of the RealBirds dataset. Better viewed with zoom-in.}
\label{birds_res}
\end{figure}

\begin{figure}[htbp]
\centering
\includegraphics[width=\textwidth]{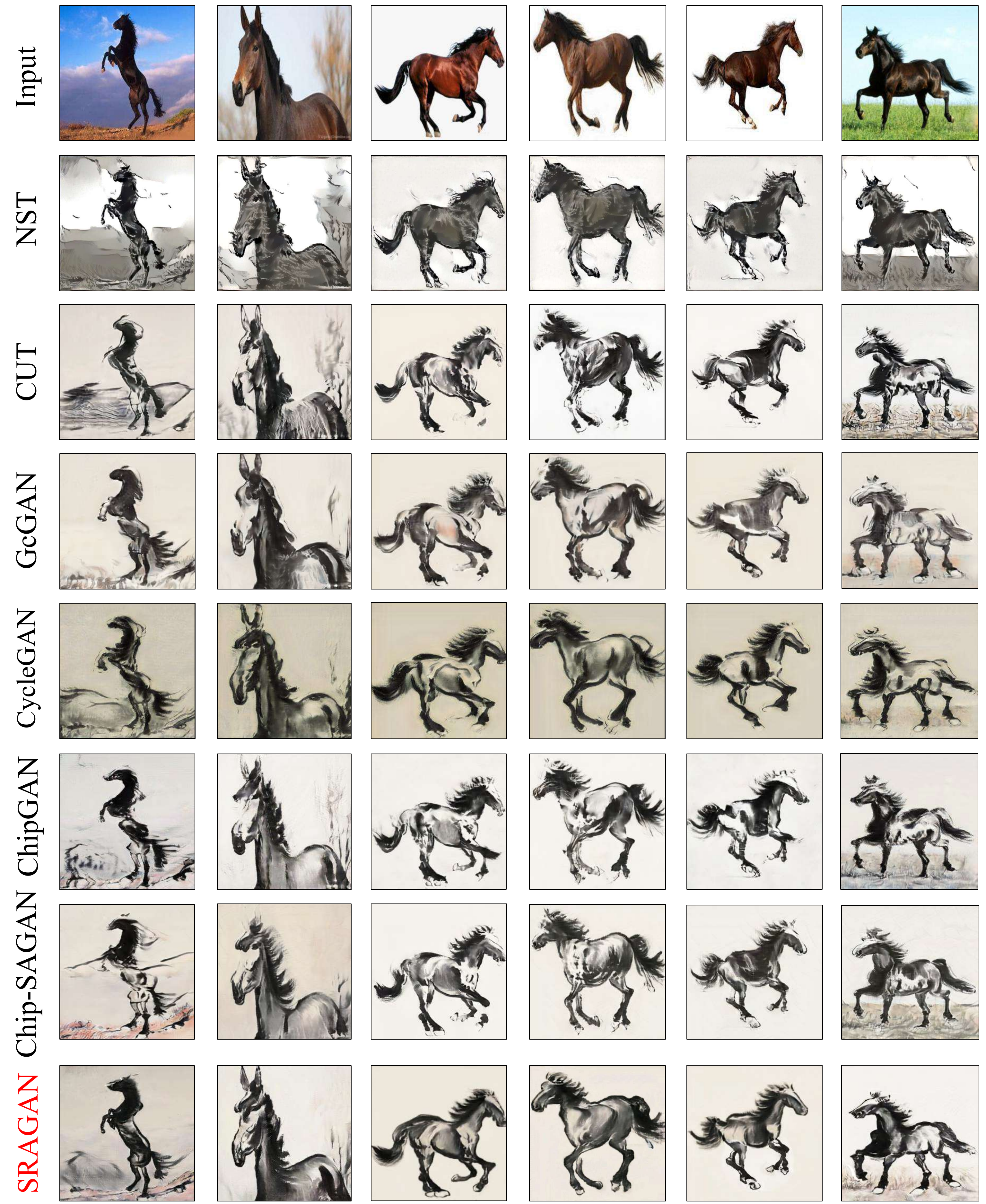}
\caption{Qualitative comparison of our SRAGAN with related GAN-based methods in Chinese ink-wash painting style transfer of horses. Our method achieves both higher content integrity and stylization quality. Bettered viewed with zoom-in.}
\label{method_compare_horse}
\end{figure}

\begin{figure}[t]
\centering
\includegraphics[width=0.9\textwidth]{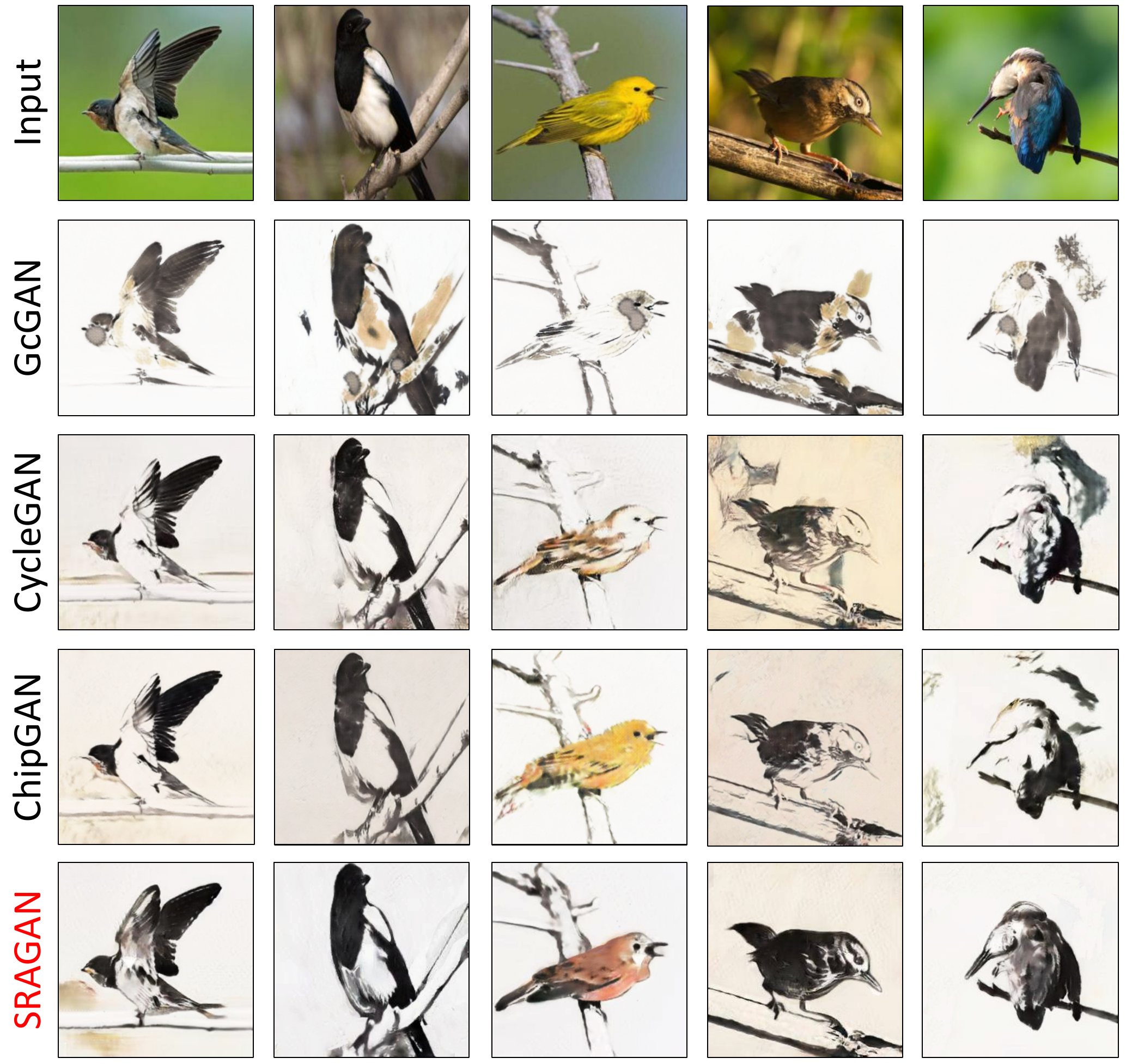}
\caption{Qualitative comparison of our SRAGAN with related GAN-based methods in Chinese ink-wash painting style transfer of birds. Our method produces results with noticeably better content integrity.}
\label{method_compare_bird}
\end{figure}

\begin{figure}[t]
\centering
\includegraphics[width=\textwidth]{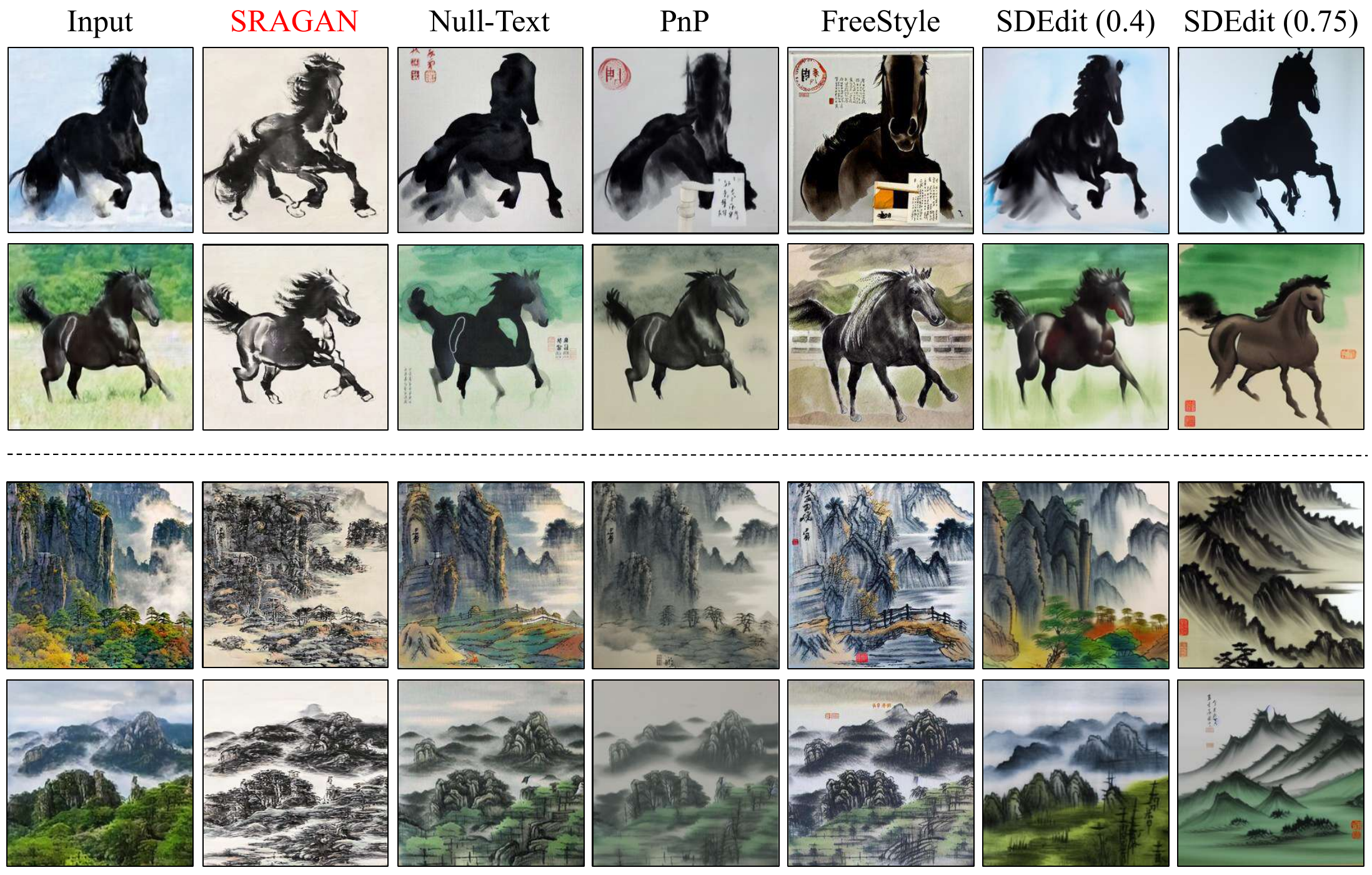}
\caption{Qualitative comparison of our SRAGAN with related diffusion-based methods in Chinese ink-wash painting style transfer of horses and landscapes. The parenthesized values denote different noising strengths in SDEdit. Better viewed with zoom-in.}
\label{compare_with_diffusion}
\end{figure}

\section{Experiments}
\subsection{Data Preparation}
We experiment on three types of typical Chinese ink-wash paintings: (\romannumeral1) landscape painting, (\romannumeral2) horse painting, (\romannumeral3) bird painting. For the landscape painting style transfer task, we use both the source-domain (RealLandscape) and target-domain (InkLandscape) datasets contributed by He B et al. \cite{bib28}. Datasets for the other two types of painting style transfer tasks are described below:
\begin{itemize}
	\item RealHorses: source-domain dataset for Chinese ink-wash painting style transfer of horses. It comprises 900 real horse pictures with diversified poses and scales collected from the Internet.
	\item InkHorses: target-domain dataset for Chinese ink-wash painting style transfer of horses. It contains 500 horse paintings drawn by some renowned Chinese ink-wash horse painters.
	\item RealBirds: source-domain dataset for Chinese ink-wash painting style transfer of birds. It comprises 900 real bird images including eagles, magpies, swallows, sparrows, orioles, etc. 
	\item InkBirds: target-domain dataset for Chinese ink-wash painting style transfer of birds. It contains 500 Chinese ink-wash bird paintings collected from the Internet, including magpies, swallows, sparrows, etc.
\end{itemize}

\subsection{Training Details}
For all the three Chinese ink-wash painting stylization tasks, the source-domain real pictures are partitioned into a training set and a test set in the proportion of 4:1. All images are resized to 256 $\times$ 256. we set $\lambda_{1}=1$, $\lambda_{2}=10$, $\lambda_{3}=5$ in Eq. \ref{total_loss}, Two Adam optimizers with momentum parameters $\beta_{1}=0.5$, $\beta_{2}=0.999$ are applied to train the generators ($G$ and $F$) and the discriminators ($D_{\mathcal{X}}$ and $D_{\mathcal{Y}}$) respectively. We train for 200 epochs with the learning rate fixed at 0.0002 in the first 100 epochs and then linearly decayed to 0 in the next 100 epochs. 

\subsection{Qualitative Evaluation}
Example qualitative evaluation results of our SRAGAN for Chinese ink-wash landscape painting, horse painting, and bird painting style transfer tasks are displayed in Fig. \ref{landscape_res}, Fig. \ref{horses_res}, and Fig. \ref{birds_res} respectively. Our method produces results with relatively precise Chinese ink-wash style, well-preserved content structure, as well as prominent visual saliency of the drawn objects.

Qualitative comparison of our method with related NST and GAN-based methods are displayed in Fig. \ref{method_compare_horse} and Fig .\ref{method_compare_bird}. Among the compared baseline methods, NST fails to learn the exact ink-wash style patterns but only transfers superficial color distribution. Results of CUT and GcGAN show noticeable artifacts and missing of content details. CycleGAN captures relatively precise style distribution but is insufficient to maintain fine object structures. ChipGAN and Chip-SAGAN improve object contour preservation by virtue of the edge prediction loss. However, they still struggle at preserving fine structures and generating prominent ink-wash elements in the internal regions inside object contours. By contrast, our method produces Chinese ink-wash paintings with more intact content structure of the drawn objects, as well as higher-quality ink-wash style rendering, thanks to our proposed saliency constrained content regularization and saliency attended generative adversarial learning. It also shows that by maximizing IOU between saliency masks rather than strictly matching saliency maps in pixel-wise manner, our method only constrains overall object structure similarity with mild contour displacement allowed, which benefits to synthesizing more vivid artistic brush strokes and more delicate ink-wash textures.

Besides, we also compare to advanced diffusion model based I2I stylization methods including Null-text inversion \cite{bib22}, PnP \cite{bib23}, FreeStyle \cite{bib24}, and SDEdit \cite{bib21}. These methods are based on pre-trained large-scale text-to-image diffusion model, for which we use Stable Diffusion v1.5 \cite{bib20} as the foundation model. To stylize an input image into a Chinese ink-wash painting, we use ``Chinese ink-wash painting of $*$'' as text prompt where $*\in\{``a \ horse", ``birds", ``landscape"\}$. Example qualitative results displayed in Fig. \ref{compare_with_diffusion} show that SDEdit with high denoising strength (e.g., SDEdit(0.75)) fails to maintain object content structure of the input image, while the counterpart with low denoising strength (e.g., SDEdit(0.4)) preserves content structure well at the expense of insufficient ink-wash style rendering. Results of Freestyle suffer from noticeable object structure distortion. In terms of stylization effect, all the compared diffusion-based methods produce results with less resemblance to the real style patterns of Chinese ink-wash paintings. By contrast, our results exhibit more precise Chinese ink-wash painting styles without sacrificing object content integrity.

\subsection{Quantitative Evaluation}

\begin{table}[t]
\begin{center}
\begin{minipage}{\textwidth}
\caption{Quantitative comparison to related NST and GAN-based methods.}\label{tab1}
\begin{tabular*}{\textwidth}{@{\extracolsep{\fill}}lcccccc@{\extracolsep{\fill}}}
\toprule
& \multicolumn{3}{@{}c@{}}{FID $\downarrow$} & \multicolumn{3}{@{}c@{}}{Saliency MIOU $\uparrow$} \\\cmidrule{2-4}\cmidrule{5-7}%
Method & Landscape & Horse & Bird & Landscape & Horse & Bird \\
\midrule
NST\cite{bib5}  & 177.43 & 186.78 & 204.95 & 0.773 & 0.790 & 0.828\\
LapStyle\cite{bib8}  & 153.54 & 179.33 & 191.98 & 0.778 & 0.805 & 0.801\\
SANet\cite{bib32} & 148.67 & 168.25 & 188.66 & 0.785 & 0.812 & 0.833\\
Adaattn\cite{bib33} & 142.86 & 165.95 & 182.79 & 0.790 & 0.814 & 0.824\\
GcGAN\cite{bib36} & 93.17 & 100.24  & 146.86  & 0.716 & 0.685 & 0.577\\
CUT\cite{bib37} & 90.56 & 98.86  & 124.67 & 0.702 & 0.696 & 0.604\\
CycleGAN\cite{bib19}  & 85.85 & 91.43 & 115.17  & 0.766 & 0.747 & 0.764\\
UNIT\cite{bib35} & 88.24 & 89.55 & 112.46 & 0.772 & 0.755 & 0.780\\
ChipGAN\cite{bib28} & 85.94 & 86.67  & 110.38 & 0.774 & 0.779 & 0.793\\
Chip-SAGAN\cite{bib29} & 86.13 & 85.45  & 111.04 & 0.777 & 0.784 & 0.792\\
SRAGAN (ours)  & \textbf{85.28} & \textbf{82.51} &  \textbf{104.96} & \textbf{0.793} & \textbf{0.820} & \textbf{0.841}\\
\bottomrule
\end{tabular*}
\footnotetext{Note: The ``Landscape'', ``Horse'', ``Bird'' denote the test set of the RealLandscape, RealHorses, RealBirds dataset, respectively.}
\end{minipage}
\end{center}
\end{table}

\begin{table}[t]
\begin{center}
\begin{minipage}{\textwidth}
\caption{Quantitative comparison to related diffusion-based methods.}\label{tab2}
\begin{tabular*}{\textwidth}{@{\extracolsep{\fill}}lcccccc@{\extracolsep{\fill}}}
\toprule
& \multicolumn{3}{@{}c@{}}{FID $\downarrow$} & \multicolumn{3}{@{}c@{}}{Saliency MIOU $\uparrow$} \\\cmidrule{2-4}\cmidrule{5-7}%
Method & Landscape & Horse & Bird & Landscape & Horse & Bird \\
\midrule
SDEdit(0.4)\cite{bib21} & 164.45 & 178.24 & 186.83 & \textbf{0.839} & \textbf{0.881} & \textbf{0.877}\\
SDEdit(0.75)\cite{bib21} & 136.34 & 152.  & 166.58  & 0.616 & 0.647 & 0.665\\
Null-Text\cite{bib22}  & 131.25 & 158.68 & 170. & 0.828 & 0.854 & 0.862\\
PnP\cite{bib23}  & 134.45 & 153.94 & 166.75 & 0.836 & 0.859 & 0.857\\
Freestyle\cite{bib24} & 129.87 & 159.86 & 170.26 & 0.770 & 0.786 & 0.818\\
SRAGAN (ours)  & \textbf{85.28} & \textbf{82.51} &  \textbf{104.96} & 0.793 & 0.820 & 0.841\\
\bottomrule
\end{tabular*}
\end{minipage}
\end{center}
\end{table}

\begin{table}[t]
\begin{center}
\begin{minipage}{\textwidth}
\caption{Quantitative ablation study on key ingredients of our method.}\label{tab3}
\begin{tabular*}{\textwidth}{@{\extracolsep{\fill}}lcccccc@{\extracolsep{\fill}}}
\toprule
& \multicolumn{3}{@{}c@{}}{FID $\downarrow$} & \multicolumn{3}{@{}c@{}}{Saliency MIOU $\uparrow$} \\\cmidrule{2-4}\cmidrule{5-7}%
Ablation & Landscape & Horse & Bird & Landscape & Horse & Bird \\
\midrule
w/o $L_{SIOU}$  & 85.62 & 86.20 & 111.35 & 0.781 & 0.781 & 0.799 \\
w/ $L_{SMSE}$  & 86.47 & 85.78 & 109.40 & \textbf{0.798} & \textbf{0.834} & \textbf{0.850} \\
w/o $SANorm$  & 85.40 & 85.54  & 108.28 & 0.789 & 0.807 & 0.832 \\
w/o $L_{SAAdv}$ & 85.35 & 84.13  & 106.23  & 0.794 & 0.831 & 0.842 \\
SRAGAN (ours) & \textbf{85.28} & \textbf{82.51}  & \textbf{104.96} & 0.793 & 0.820 & 0.841 \\
\bottomrule
\end{tabular*}
\end{minipage}
\end{center}
\end{table}

For quantitative evaluations, we use FID to measure the domain distribution distance between image set of real Chinese ink-wash paintings and the generated ink-wash paintings. The FID value between generated data distribution $P_{g}$ and real data distribution $P_{r}$ is defined as:
\begin{equation}
FID(P_{r}, P_{g})=\|\mu_{r}-\mu_{g}\|+T_{r}(\sigma_{r}+\sigma_{g}-2(\sigma_{r}\sigma_{g})^{1/2}),
\end{equation}
where $\mu_{r}$ and $\sigma_{r}$ are the mean and variance of real data features extracted by the pre-trained Inception V3 \cite{bib38} model, $\mu_{g}$ and $\sigma_{g}$ are the mean and variance of the features extracted from the generated data. The lower the FID is, the more similar the generated Chinese ink-wash paintings are to the real ones. Besides, we also use the saliency detection and thresholding pipeline to extract saliency masks for both real-world input pictures and the stylized ink-wash paintings, and calculate the mean IOU (which we denote as Saliency MIOU) between the extracted saliency masks of the two domains to measure object content integrity after ink-wash painting stylization:
\begin{equation}
Saliency \ MIOU(\mathcal{X})= \frac{1}{N}\sum_{x \in \mathcal{X}}IOU[T(S(x)), T(S(G(x)))],
\end{equation}
where $N$ is the number of test samples in $\mathcal{X}$. Quantitative comparison of our method with related NST and GAN-based methods are reported in Tab. \ref{tab1}. It shows that NST-based methods \cite{bib5}\cite{bib8}\cite{bib32}\cite{bib33} perform worse in FID than GAN-based methods since they model image style as low-level textures, not sufficient to capture the precise high-level ink-wash style patterns. In comparison to the compared GAN-based methods, our approach achieves the lowest FID as well as the highest Saliency MIOU in all three tasks, indicating the superiority of our model in style capture accuracy and content preservation capability, thanks to our proposed saliency-guided style enhancement and content regularization. When compared with diffusion-based approaches, evaluation results reported in Tab. \ref{tab2} show that our method outperforms related advanced text-to-image diffusion models by a large margin in FID, indicating that the style patterns learned by our method are closer to real Chinese ink-wash paintings compared to the prior knowledge of large-scale text-to-image diffusion models learned from massive open-domain image data.

\begin{figure}[t]
\centering
\includegraphics[width=\textwidth]{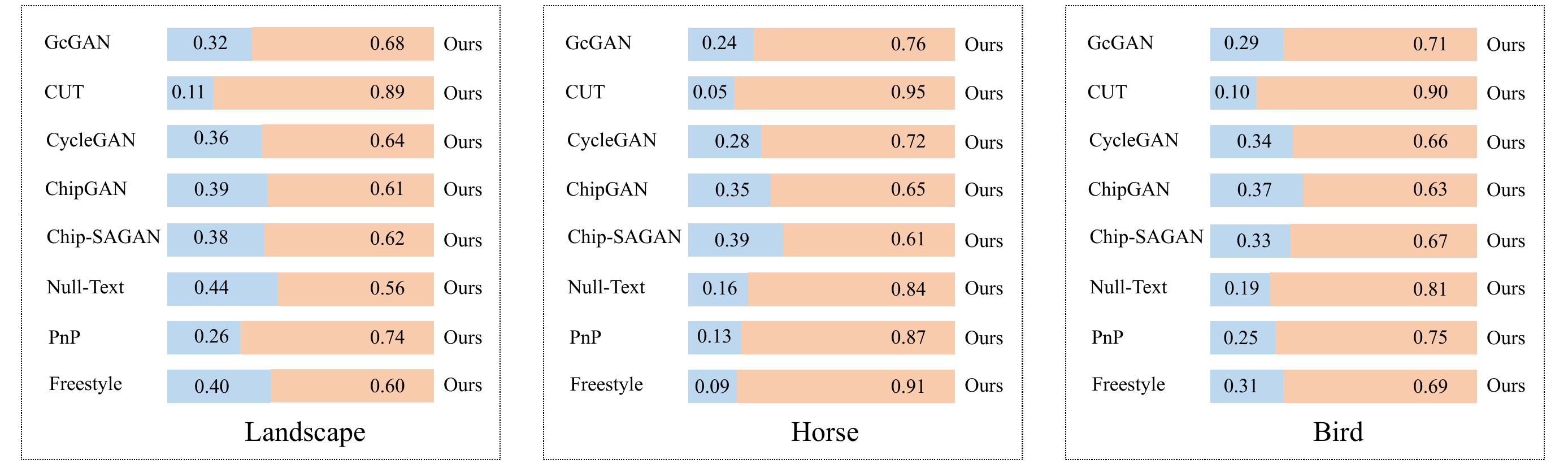}
\caption{Study on the user preference to our approach against related methods.}
\label{user_study}
\end{figure}

Besides, user study is conducted to evaluate methods from subjective perspective. For all the three Chinese ink-wash painting style transfer tasks, we randomly sample 100 pictures from the corresponding test set to evaluate our model and related competitive methods. For each task, 34 participants are invited to judge whether results of our method are more similar to real Chinese ink-wash paintings than each compared method. The proportion of user preference to our approach against related methods on different Chinese ink-wash painting style transfer tasks are reported in Fig. \ref{user_study}. Our method gains more user preference votes than all the compared methods including both GAN-based and diffusion-based ones in all the three tasks, reflecting that our results more visually resemble real Chinese ink-wash paintings from subjective perspective.

\begin{figure}[htbp]
\centering
\includegraphics[width=\textwidth]{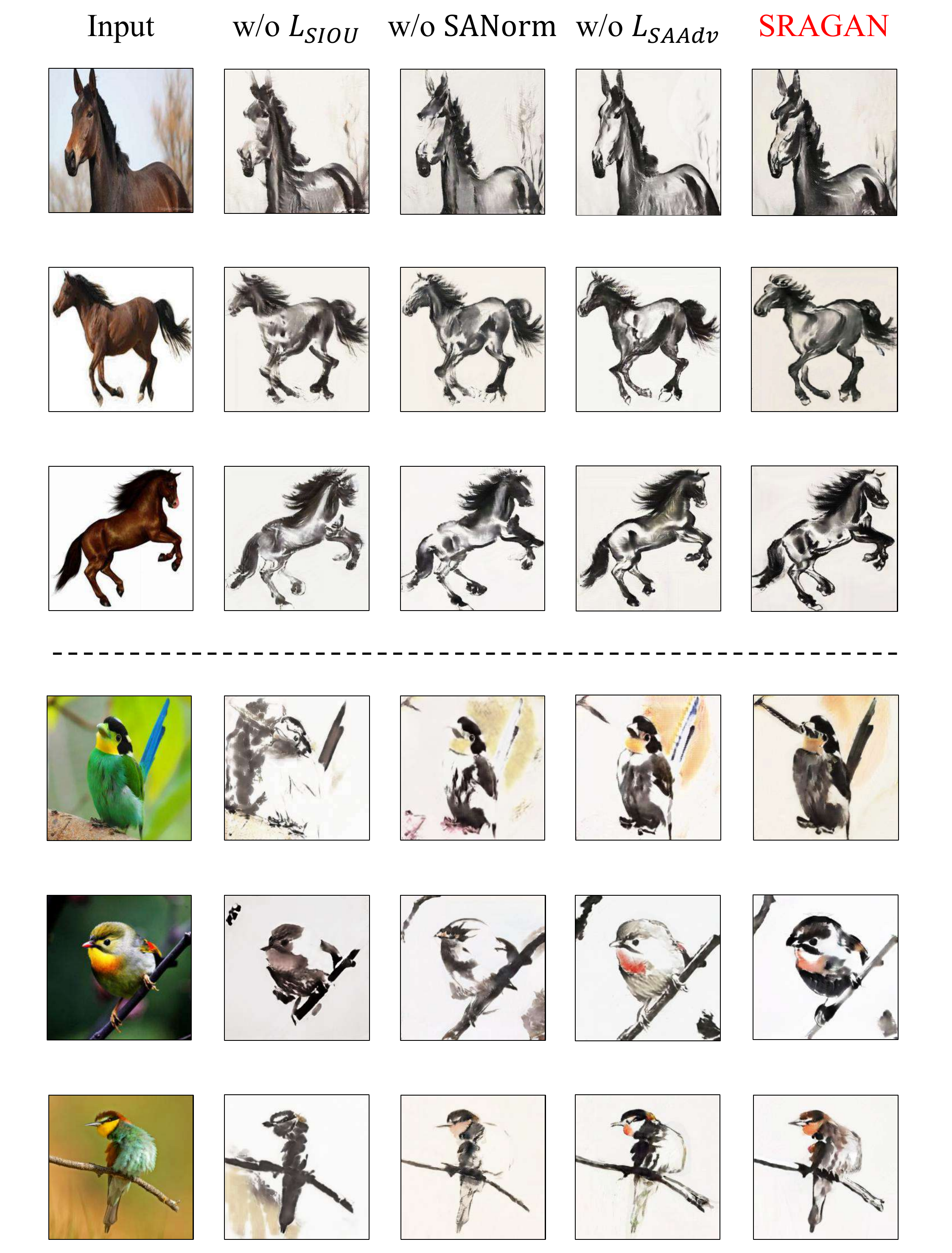}
\caption{Qualitative ablation study on the key ingredients of our model, including the saliency IOU loss $L_{SIOU}$, saliency adaptive normalization (SANorm), and the saliency attended adversarial loss $L_{SAAdv}$.}
\label{ablation_study}
\end{figure}

\begin{figure}[t]
\centering
\includegraphics[width=\textwidth]{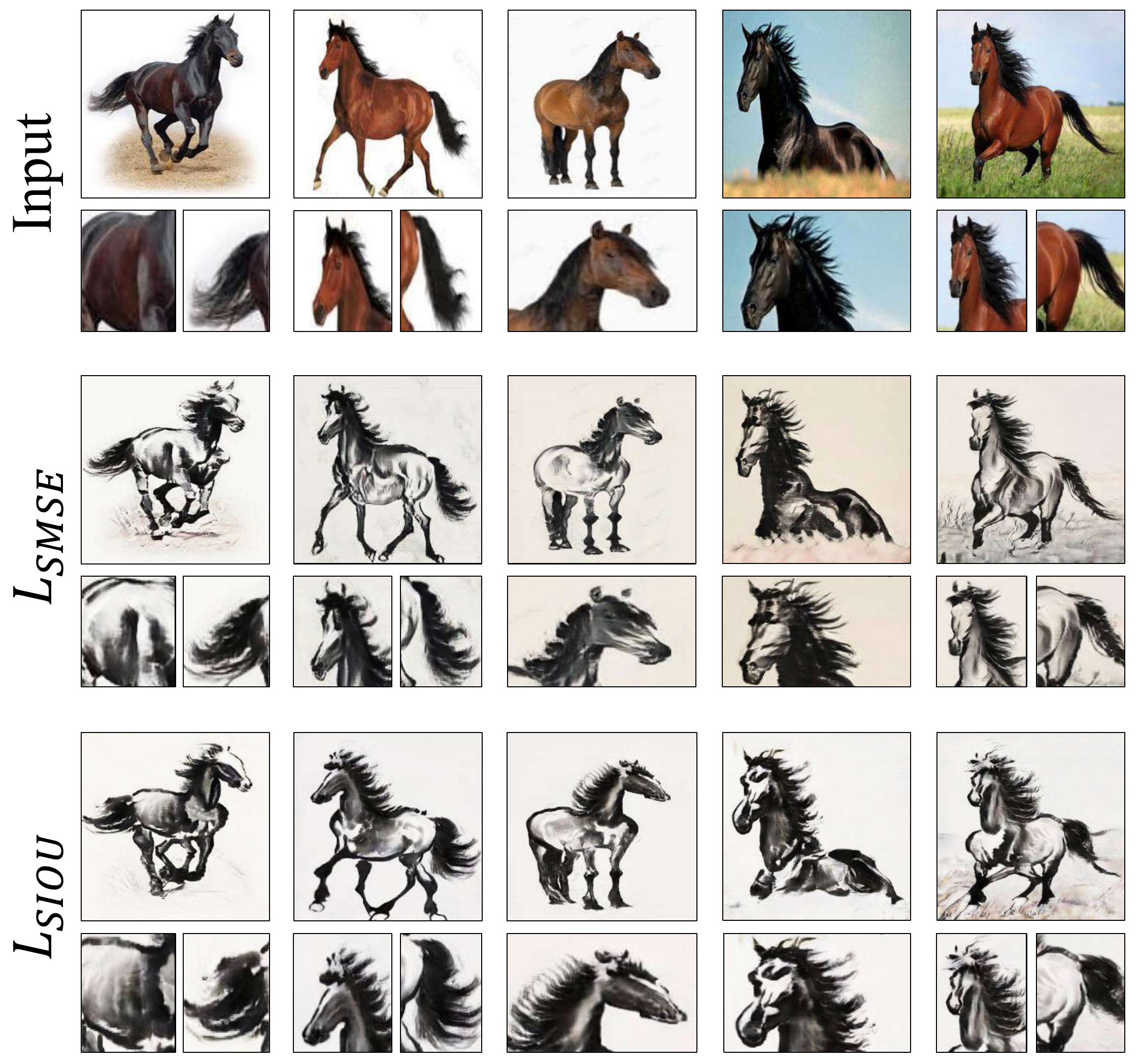}
\caption{Qualitative comparison between the pixel-wise saliency MSE loss $L_{SMSE}$ and our proposed saliency IOU loss $L_{SIOU}$. Better viewed with zoom-in.}
\label{saliency_loss_study}
\end{figure}

\subsection{Ablation Study}
To investigate the effectiveness of the key ingredients of our method, we design ablation studies with notations explained as follows:
\begin{itemize}
	\item w/o $L_{SIOU}$: the model without saliency IOU loss.
	\item w/o SANorm: the model without SANorm layers, i.e., all SNblocks in the generator are replaced with normal Resblocks.
	\item w/o $L_{SAAdv}$: the model that replaces our saliency attended adversarial loss with normal MSE-based adversarial loss, and replaces our saliency attended discriminator with normal PatchGAN discriminator \cite{bib18}.
	\item w/ $L_{SMSE}$: the model that replaces our saliency IOU loss $L_{SIOU}$ with saliency MSE loss $L_{SMSE}$, the pixel-wise mean squared error between the detected saliency maps before and after image stylization.
\end{itemize}

As shown in Fig. \ref{ablation_study}, the absence of $L_{SIOU}$ leads to the stylized objects with noticeable missing of content details. The same issue also appears for the case of w/o SANorm. These reflect the importance of our explicit and implicit saliency regularization in improving image stylization content integrity. Comparing results of w/o $L_{SAAdv}$ to that of the full model, it shows that our proposed saliency attended adversarial learning contributes to generating more vivid and delicate ink-wash brush strokes for the drawn objects. By combining both saliency constrained content regularization and saliency attended adversarial learning, the ultimate full model can balance content integrity and style faithfulness satisfactorily. 

In Fig. \ref{saliency_loss_study}, we qualitatively demonstrate superiority of our proposed saliency IOU loss $L_{SIOU}$ over directly minimizing pixel-wise MSE between the detected saliency maps, i.e., $L_{SMSE}$. It shows that $L_{SMSE}$ is too strict to produce vivid artistic brush strokes, the stylization results are relatively rigid. Our saliency IOU loss, by contrast, relaxes structural constraint by enforcing only overall object structure similarity rather than pixel-wise alignment, which benefits to producing more abstract and vivid artistic brush strokes and ink-wash textures to mimic the specific drawing skills of Chinese ink-wash painting.

\begin{figure}[t]
\centering
\includegraphics[width=0.9\textwidth]{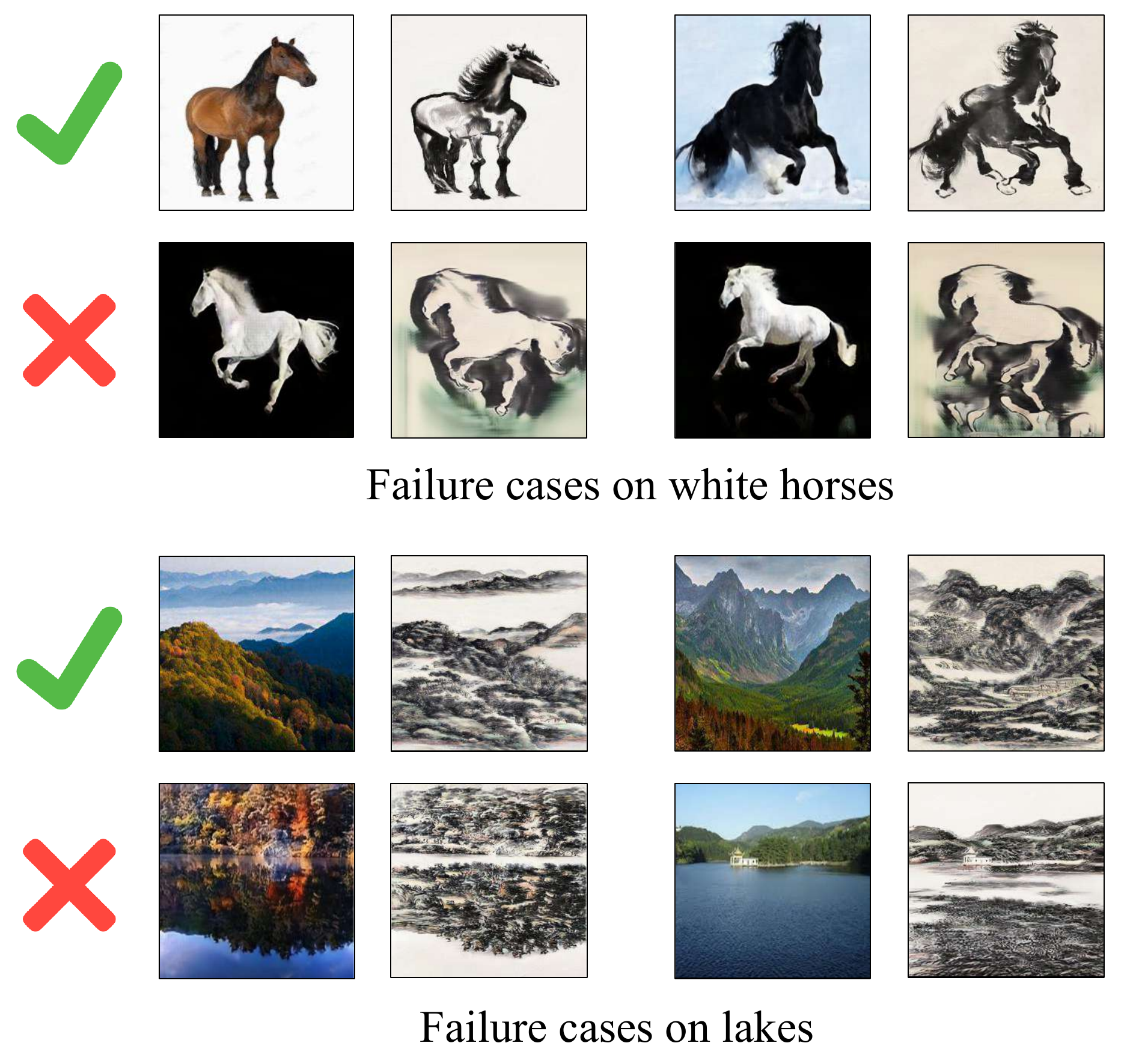}
\caption{Failure cases of our method on stylization of white horses and landscape pictures with lakes.}
\label{failure_case}
\end{figure}

We also conduct quantitative ablation study in terms of FID and Saliency MIOU. Results displayed in Tab. \ref{tab3} show that both the saliency IOU loss and the SANorm have evident contributions to promoting Saliency MIOU, i.e., improving content integrity. These two ingredients also improve FID due to their effectiveness in alleviating object content corruption issue. The introduction of saliency attended adversarial learning brings consistent improvements in FID at the cost of slight drops in Saliency MIOU. Compared with saliency IOU loss, saliency MSE loss leads to enhanced structure consistency, while causing worse FID since pixel-wise saliency constraint confines model's ability to synthesize stylistically faithful ink-wash textures and vivid brush strokes.

\section{Limitation and Discussion}
There are also failure cases in our results due to less diversified training data and semantic unawareness of input images, which we showcase in Fig. \ref{failure_case}. In the track of Chinese ink-wash horse painting style transfer, our method produces fine results for horses with dark colors but fails in the case of white horses. This could be due to the rareness of white horses in the training data such that the model is overfitted to dark horses. In the track of Chinese ink-wash landscape painting style transfer, our method fails to generalize well to landscape pictures with lakes, since our current model cannot distinguish content semantics and stylize content in a semantic-aware manner. In future work, we may tackle these problems from three aspects: (\romannumeral1) increase training data diversity to alleviate long tail issue; (\romannumeral2) introduce clustering algorithm to discover latent clusters of training data and use cluster prototypes as conditional signals to guide the network for prototype-agnostic style transfer; (\romannumeral3) leverage cutting-edge open-domain semantic grounding model to embed semantic priors into our framework. Besides, inspired by recent works of learning Positive-Incentive Noise (Pi-Noise) to boost base model performance \cite{bib39}\cite{bib40}, we will investigate how to transfer the Pi-Noise framework to the GAN-based generative model, improving image generation and I2I stylization visual quality without changing base model architecture. Last but not least, we also plan to significantly increase the resolution of the generated paintings by upgrading the network to larger capacity or training a conditional upscaling model for postprocessing.

\section{Conclusion}
Aimed to leverage saliency detection to enhance object structure integrity and promote stylization visual quality for the task of Chinese ink-wash painting style transfer, this paper proposes SRAGAN, a GAN-based unsupervised I2I stylization framework that applys saliency detection to assist image stylization from three aspects. Firstly, we propose saliency IOU loss that explicitly regularizes object content structure from the optimization function perspective, dramatically alleviating object content corruption issue caused by transfer of irregular ink-wash style elements. Secondly, we propose saliency regularized generator that implicitly regularizes content structure of the generated paintings through dynamic image saliency guidance. By fusing structural information from image saliency map into network forward propagation via our proposed saliency adaptive normalization (SANorm), the object structure integrity of the translated paintings is further improved. Thirdly, we propose saliency attended generative adversarial learning which employs the detected saliency map to adaptively focus adversarial stylization attention onto salient image objects, contributing to synthesizing more abstract, delicate, and vivid brush strokes and ink-wash textures. Integrating the above-mentioned contributions together, our model exhibits noticeable advantages over related NST, GAN, and diffusion based methods in Chinese ink-wash painting style transfer, both in object structure integrity and target style faithfulness. The major weakness of our model is that it is established on GAN-based I2I framework, which is relatively old compared to the currently prevailing diffusion-based paradigm. Nevertheless, our method pioneer combining saliency detection with generative model for artistic style transfer, contributing a novel solution to boost Chinese ink-wash painting stylization in both content and style, as well as a new perspective to assist generative AI with traditional computer vision problem. Unvailing that saliency detection deep model pre-trained on real-world images also generalize to artistic domain, our work broadens the application scope of image saliency detection and may inspire its potential applications in more AI art creation scenarios.

\end{document}